\documentclass[letterpaper, 10 pt, conference]{ieeeconf}  

\IEEEoverridecommandlockouts                              

\usepackage{amsmath} 
\usepackage{amssymb}  
\usepackage{todonotes}
\let\labelindent\relax 
\usepackage{enumitem}
\usepackage{mathtools}
\usepackage{rotating}
\usepackage{multirow}
\usepackage{siunitx}
\usepackage{caption}
\usepackage{subcaption}
\usepackage{float}
\usepackage{hyperref}
\usepackage{pifont}
\usepackage{cite}
\newcommand{\xmark}{\ding{55}}%
\usepackage{microtype}

\usepackage{booktabs} 

\usepackage{algorithmic} 
\usepackage{algorithm2e}
\graphicspath{{graphics/}}

\usepackage{tabto} 

\hypersetup{
    colorlinks=true,
    linkcolor=black,
    filecolor=magenta,      
    urlcolor=cyan,
    pdftitle={Overleaf Example},
    pdfpagemode=FullScreen,
    }

\title{\LARGE \bf
SCOPE: A Synthetic Multi-Modal Dataset for Collective Perception Including Physical-Correct Weather Conditions
}

\author{Jörg Gamerdinger$^{1}$, Sven Teufel$^{1}$, Patrick Schulz$^{2}$, Stephan Amann$^{1}$, \\Jan-Patrick Kirchner$^{1}$,  and Oliver Bringmann$^{1}$
\thanks{$^{1}$University of T\"ubingen, Faculty of Science, Department of Computer Science, Embedded Systems Group 
\tt\small {\{joerg.gamerdinger, sven.teufel, stephan.amann, jan-patrick.kirchner, oliver.bringmann\} @uni-tuebingen.de}}
\thanks{$^{2}$ Forschungszentrum Informatik (FZI) Karlsruhe 
\tt\small {schulz@fzi.de}}
}%

\begin{document}
\maketitle
\thispagestyle{empty}
\pagestyle{empty}

\begin{abstract}

Collective perception has received considerable attention as a promising approach to overcome occlusions and limited sensing ranges of vehicle-local perception in autonomous driving. In order to develop and test novel collective perception technologies, appropriate datasets are required. These datasets must include not only different environmental conditions, as they strongly influence the perception capabilities, but also a wide range of scenarios with different road users as well as realistic sensor models. Therefore, we propose the Synthetic COllective PErception (SCOPE) dataset. SCOPE is the first synthetic multi-modal dataset that incorporates realistic camera and LiDAR models as well as parameterized and physically accurate weather simulations for both sensor types. The dataset contains 17,600 frames from over 40 diverse scenarios with up to 24 collaborative agents, infrastructure sensors, and passive traffic, including cyclists and pedestrians. In addition, recordings from two novel digital-twin maps from Karlsruhe and Tübingen are included.
\ \\
The dataset is available at \url{https://ekut-es.github.io/scope}

\end{abstract}


\section{INTRODUCTION}
\label{sec:intro}

A comprehensive perception of the environment is crucial for the safe operation of autonomous vehicles. However, vehicle-local perception is limited by sensing ranges and occlusions, and is also affected by environmental conditions such as rain, snow, and fog~\cite{volk2019towards,Teufel-IV23}. Collective Perception (CP) is a promising approach to overcome these challenges by significantly increasing perception range, performance, and safety~\cite{volk_environment-aware_2019,schiegg2021,gamerdinger2023cold,teufel2023collective}.

There are several well-known datasets in autonomous driving research, such as KITTI~\cite{kitti} or WAYMO~\cite{waymo}. However, these datasets consist of recordings from a single vehicle and, as a result, are not suitable for training and testing collective perception algorithms.  
Currently available CP datasets lack scenario diversity, realistic sensor models, environmental conditions, or vulnerable road users (VRUs)~\cite{teufel2024review}.
Therefore, we propose the novel \textbf{S}ynthetic \textbf{CO}llective \textbf{PE}rception (SCOPE) dataset. SCOPE contains recordings from over 40 different scenarios, including edge cases such as tunnels and roundabouts with up to 20 connected and automated vehicles (CAVs), roadside units (RSUs), and passive traffic including cars, vans, motorcycles, cyclists and pedestrians. In total, SCOPE consists of 17,600 frames. In addition, we have incorporated realistic LiDAR models from~\cite{rosenberger2020sequential} to address the limitations of the commonly used CARLA~\cite{CARLA} LiDAR sensor, resulting in a more authentic dataset. Moreover, SCOPE is the only dataset to include realistic weather simulations with known intensities for both camera and LiDAR data.

\begin{figure}[t!]
    \centering
    \includegraphics[width=\linewidth, page=2, trim= 10cm 2.7cm 10cm 2.8cm, clip]{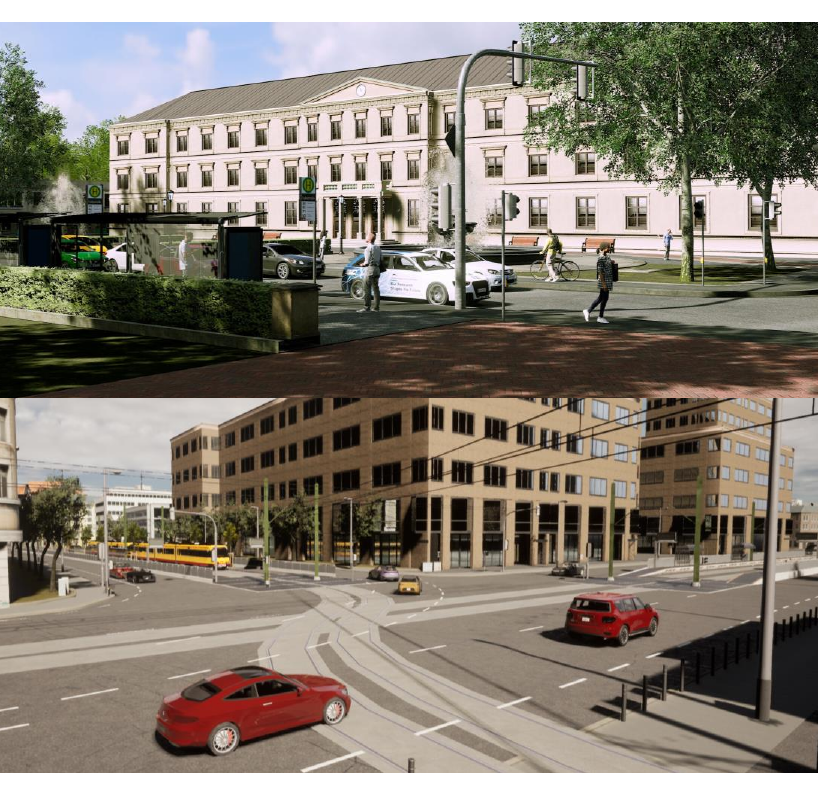}
    \caption{Exemplary scenes from Tübingen (top) and Karlsruhe (bottom)}
    \label{fig:own_maps}
\end{figure}

Our main contributions are:
\begin{itemize}
    \item SCOPE is the first synthetic multi-modal CP dataset with realistic LiDAR sensor models and solid-state LiDARs
    \item We present the first CP dataset with physically-accurate and parameterized weather simulations for cameras and LiDARs
    \item SCOPE is the first synthetic dataset for CP that contains both pedestrians and two-wheeled vehicles
    \item We include two novel maps from Karlsruhe and Tübingen, to obtain a more diverse dataset
\end{itemize}

In Section~\ref{sec:related_work} we provide an overview about currently available collective perception datasets and discuss their limitations. The SCOPE dataset is presented in detail in Section~\ref{sec:method}. We then provide, details about the object detection and semantic segmentation benchmark. 
Finally, in Section~\ref{sec:conclusion} we conclude our work and give an outlook on further extensions of the dataset.
\begin{table*}
\centering
\caption{Overview of synthetic collective perception datasets. Adapted from~\cite{teufel2024review}}
\renewcommand{\arraystretch}{1.0}
\resizebox{\textwidth}{!}{%
\begin{tabular}{lllllllllllll}
\hline
Dataset &  CAVs & RSUs & Frames & LiDAR & \begin{tabular}[c]{@{}l@{}}Realistic \\ LiDAR\end{tabular} & RGB Camera & SemSeg &Labels** & Classes & Scenarios & Weather & \begin{tabular}[c]{@{}l@{}}Directly \\ Available\end{tabular} \\ \hline
V2V-Sim \cite{wang2020v2vnet} \begin{tabular}[c]{@{}l@{}}\ \\ \ \end{tabular} &  $\approx$10 & no & 51k** & ? & \xmark & - &\xmark & ? & Car & ? &\xmark & \xmark \\ \hline
CODD \cite{codd}& 4-16 & no & 13k & 64 Layer & \xmark& - & \xmark& 204k & Car, Pedestrian & \begin{tabular}[c]{@{}l@{}}urban, suburban, \\ rural\end{tabular} & \xmark  & \checkmark \\ \hline
COMAP \cite{comap} \begin{tabular}[c]{@{}l@{}}\ \\ \ \end{tabular}&  2-20 & no & 8.6k & 32 Layer & \xmark& 800$\times$600\,px* & \checkmark &226k & Car, Truck & urban, suburban & \xmark  & \checkmark \\ \hline
V2XSet \cite{xu2022v2xvit} &  2-7 & yes & 11k & 32 Layer &  \xmark& - & \xmark &254k & Car & \begin{tabular}[c]{@{}l@{}}urban, suburban, \\ rural, highway\end{tabular} & \xmark &  \checkmark \\ \hline
DOLPHINS \cite{dolphins} &  2 & yes & 42k & \begin{tabular}[c]{@{}l@{}}64 Layer \\ (RSU\&CAV)\end{tabular} & \xmark&  \begin{tabular}[c]{@{}l@{}}1920$\times$1080\,px (RSU)\\ 1920$\times$1080\,px (CAV)\end{tabular} & \xmark &292k & Car, Pedestrian & \begin{tabular}[c]{@{}l@{}}urban, suburban, \\ rural, highway\end{tabular} & \xmark & \xmark \\ \hline
OPV2V \cite{opv2v} &  2-7 & no & 11k & 64 Layer & \xmark& 4x 800$\times$600\,px & \checkmark &232k & Car & \begin{tabular}[c]{@{}l@{}}urban, suburban, \\ rural, highway\end{tabular} & \xmark & \checkmark \\ \hline
V2X-Sim 2.0 \cite{V2XSim} &  2-5 & yes & 10k & \begin{tabular}[c]{@{}l@{}}32 Layer \\ (RSU\&CAV)\end{tabular} & \xmark& \begin{tabular}[c]{@{}l@{}}4x 1600$\times$900\,px (RSU)\\ 6x 1600$\times$900\,px (CAV)\end{tabular} & \checkmark &4.2M & \begin{tabular}[c]{@{}l@{}}Car, Cyclist, \\ Motorcyclist\end{tabular} & \begin{tabular}[c]{@{}l@{}}urban, suburban, \\ rural, highway\end{tabular} & \xmark  & \checkmark \\ \hline
IRV2V \cite{wei2023asynchrony} \begin{tabular}[c]{@{}l@{}}\ \\ \ \end{tabular}&  2-5 & no & 8.4k & 32 Layer & \xmark& 4x 600$\times$800\,px & \xmark &1.5M & ? & ? & \xmark  & \xmark \\ \hline\hline
\textbf{SCOPE (ours)} & 3-21 & yes & 17k & \begin{tabular}[c]{@{}l@{}}64 Layer (\SI{360}{\degree)}\\32 Layer (\SI{360}{\degree)}\\ 32 Lines Solid State \\ (RSU\&CAV)\end{tabular} & \checkmark& \begin{tabular}[c]{@{}l@{}}2-4x 1920$\times$1080\,px (RSU)\\ 5x 1920$\times$1080\,px (CAV)\end{tabular} &\checkmark & 575 k & \begin{tabular}[c]{@{}l@{}}Car, Cyclist\\ Motorcyclist\\ Pedestrian, Van\end{tabular} & \begin{tabular}[c]{@{}l@{}}urban, suburban, \\ rural, highway\\Karlsruhe, Tübingen\end{tabular} & \checkmark  & \checkmark \\ \hline
\end{tabular}%
}
\caption*{\small- not present in the dataset, * recorded, but currently not included, **may contain duplicates, ? unknown}
\vspace{-0.1cm}
\label{tab:overview}
\end{table*}
\newpage
\section{RELATED WORK}
\label{sec:related_work}

Several synthetic datasets on collective perception already exist, each with its own strengths and weaknesses. An overview is given in Tab.~\ref{tab:overview}.  Due to the domain gap between real-world and synthetic datasets, we will only focus on synthetic datasets in this section.
For a comprehensive review on collective perception datasets, including real-world datasets and infrastructure-only datasets, we refer to Teufel et al.~\cite{teufel2024review}.

The first synthetic CP dataset was V2V-Sim~\cite{wang2020v2vnet} in 2020 which consists of 51k frames captured from up to 63 CAVs. It it the only synthetic dataset created using LiDARsim~\cite{manivasagam2020lidarsim} instead of CARLA~\cite{CARLA}. However, the dataset is not publicly available, and there is no information available regarding the sensor configuration or scenarios. The dataset also lacks VRUs. The CODD dataset~\cite{codd} is a pure LiDAR dataset with 4-16 CAVs which also includes pedestrians. With only 13k frames the dataset is rather small and it does not provide highway scenarios and no passive traffic resulting to less realistic scenarios. 

The COMAP dataset~\cite{comap} also has a wide range of CAVs (2-20), but with 8.6k frames it is not sufficient for comprehensive training and evaluation of neural networks. One advantage of the dataset is that it includes RGB and semantic segmentation (SemSeg) cameras. However, the images from these cameras are not yet available. Furthermore, COMAP consists only of intersection scenarios on a single map which makes it unsuitable for a comprehensive evaluation.
The V2XSet~\cite{xu2022v2xvit} from 2022 consists of 2-7 CAVs equipped with one 32-layer LiDAR sensor each. With 55 scenarios on 8 maps, the dataset has a high scenario variation. In addition, the dataset includes roadside units (RSUs). However, the dataset lacks VRUs, which limits its applicability. Moreover, the dataset has some anomalies with unrealistic vehicle positioning and accidents.
The DOLPHINS dataset~\cite{dolphins} has a sufficient size with 42k frames and shows a wide range of scenarios including urban, suburban, rural, and highway scenarios. The CAVs are equipped with a 64-layer LiDAR and a RGB camera, making this dataset suitable for LiDAR and camera-based object detection. Furthermore, this dataset includes a RSU. However, the dataset only contains up to 3 collaborative agents (vehicles or RSUs), which makes it unsuitable for experiments with varying V2X equipment rates. In addition, the dataset is currently not publicly available.

The OPV2V dataset~\cite{opv2v} is similar in scenarios and number of CAVs to the V2XSet dataset, as both use the OpenCDA~\cite{xu2021opencda} platform for the generation. The dataset also contains 2-7 CAVs, but in this dataset they are equipped with a 64-layer LiDAR and 4 RGB cameras. The number of frames and labels as well as the scenario variation are nearly the same as for V2XSet. Like the V2XSet, the dataset lacks VRUs and has only a small range of CAVs and no RSUs.

V2X-Sim~\cite{V2XSim} was released in 2022 and contains 2-5 CAVs and RSUs. Both are equipped with a 32-layer LiDAR and 1600$\times$900 px RGB cameras with \SI{360}{\degree} sensor coverage. This allows the evaluation of LiDAR-camera fusion methods. However, the dataset consists only of intersection scenarios, which leads to a low scenario diversity. Furthermore, the number of CAVs is rather small.
A dataset with temporal asynchronies is the IRV2V dataset~\cite{wei2023asynchrony}. This dataset contains 2-5 CAVs equipped with 4 800$\times$600 px RGB cameras and a 32-layer LiDAR. The scenarios in the dataset contain an average of 48 vehicles, which results in a low V2X equipment rate. Furthermore, the dataset is not yet publicly available, and various information, such as the diversity of scenarios and the object classes included, are not known.

\section{SCOPE DATASET}
\label{sec:method}


\subsection{Simulation Setup}
\label{subsec:generation}
CARLA~\cite{CARLA} is a widely used simulation environment for prototyping autonomous driving. It allows the spawning and control of different types of road users and the equipping of these users with a wide range of sensor types. Furthermore, CARLA provides a variety of maps, including controllable times of day, which can be used to simulate different and realistic environments. In the context of traffic control, CARLA can be employed in a co-simulation environment with SUMO~\cite{SUMO}, a widely used traffic simulator that facilitates the efficient generation of traffic flows. 
However, the CARLA-SUMO co-simulation in CARLA 0.9.14 is unable to spawn and control pedestrians. Therefore, we utilize the CARLA traffic manager for both vehicles and pedestrians to avoid interferences between the traffic management systems.
To generate time-aligned sensor data, we employ the use of snapshots of the world, which are then formatted and written to HDF5 containers.

The RESIST framework by Müller et al~\cite{resist} is employed for weather simulation on the images and the data dumping from HDF5 to an easily usable data structure. The framework allows for a flexible configuration of data processing and augmentation with different rain and fog intensities. Finally, the LiDAR point cloud weather simulation is applied with the framework developed by Teufel et al.~\cite{lidar-weather}.

\subsection{Scenarios}
\label{subsec:scenarios}
\begin{figure*}
     \centering
     \begin{subfigure}[b]{0.3\textwidth}
         \centering
         \includegraphics[width=\textwidth, trim= 1cm 2cm 1cm 0cm]{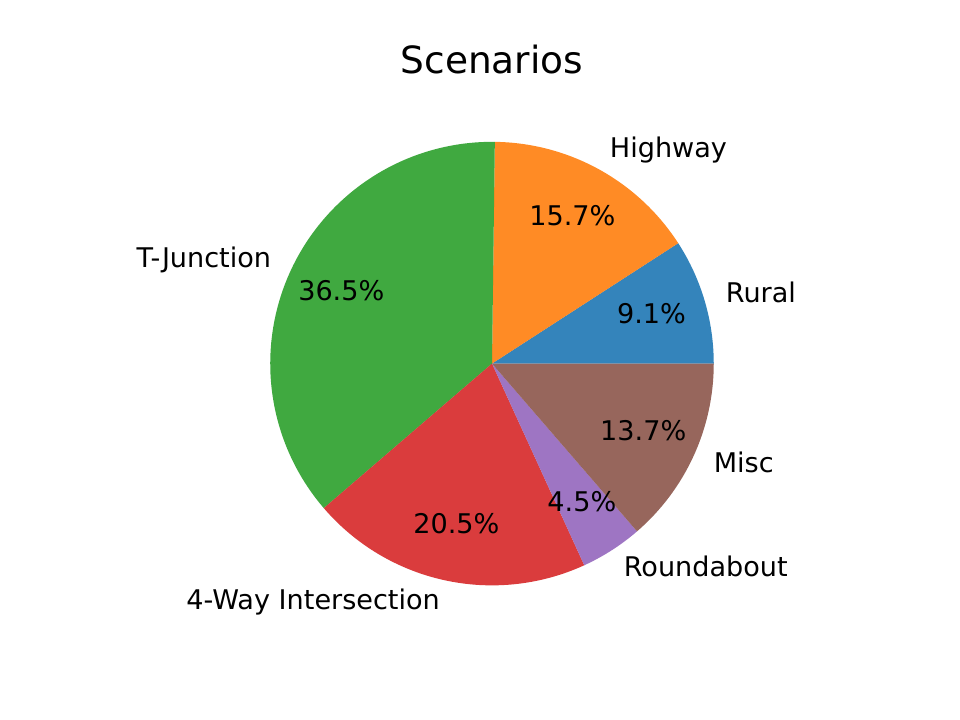}
         \caption{Scenario Distribution}
         \label{fig:scenario_distribution}
     \end{subfigure}
     \begin{subfigure}[b]{0.3\textwidth}
         \centering
         \includegraphics[width=\textwidth, trim= 1cm 2cm 1cm 0cm]{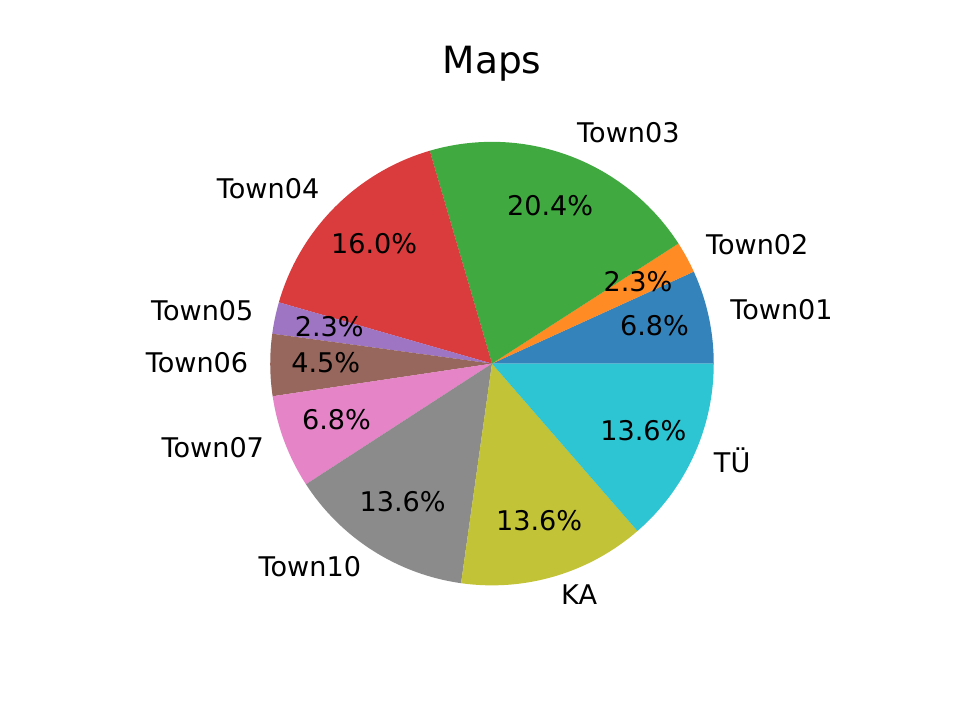}
         \caption{Map Distribution}
         \label{fig:map_distribution}
     \end{subfigure}
     \begin{subfigure}[b]{0.3\textwidth}
         \centering
         \includegraphics[width=\textwidth, trim= 1cm 2cm 1cm 0cm]{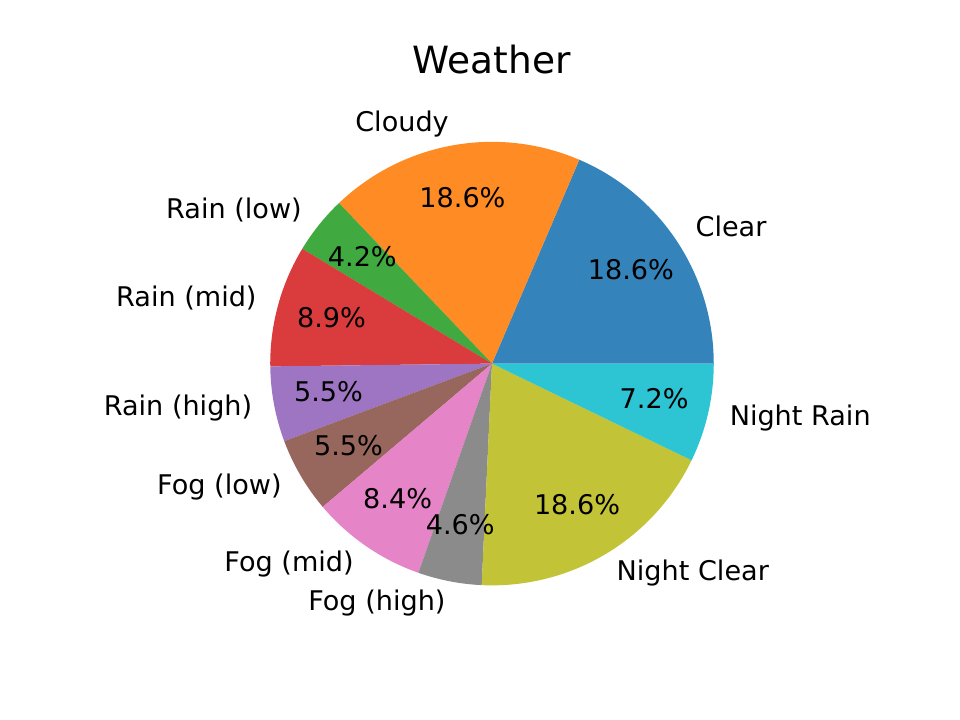}
         \caption{Weather Distribution}
         \label{fig:weather_distribution}
     \end{subfigure}
     \begin{subfigure}[b]{0.45\textwidth}
         \centering
         \includegraphics[width=.8\textwidth, trim= 1cm 0cm 1cm 0cm]{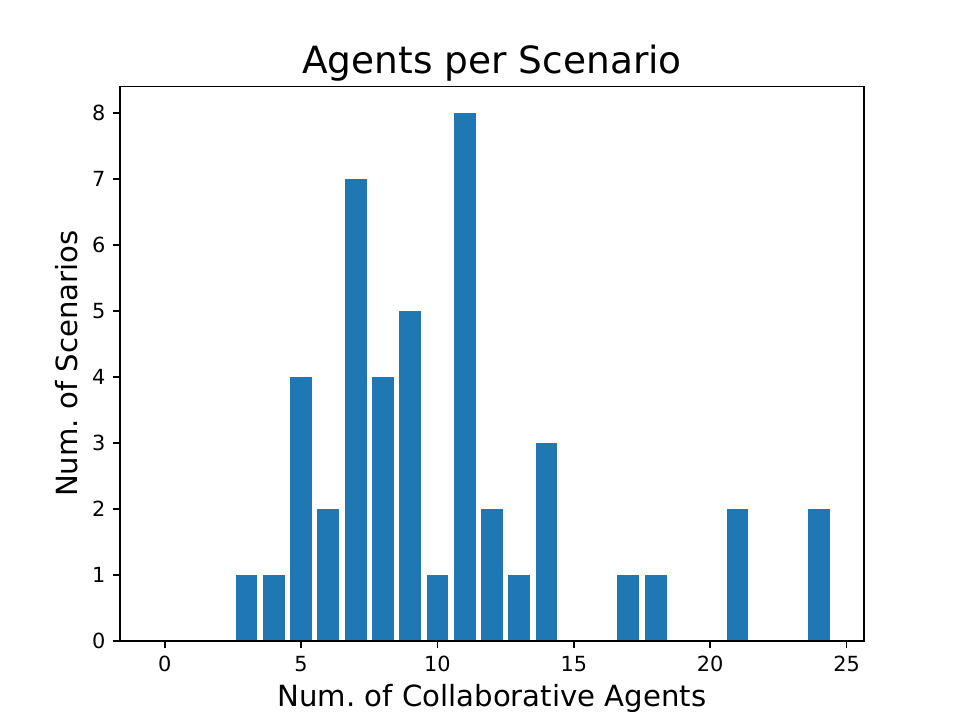}
         \caption{Agent Distribution}
         \label{fig:agent_distribution}
     \end{subfigure}
     \begin{subfigure}[b]{0.45\textwidth}
         \centering
         \includegraphics[width=.8\textwidth, trim= 1cm 0cm 1cm 0cm]{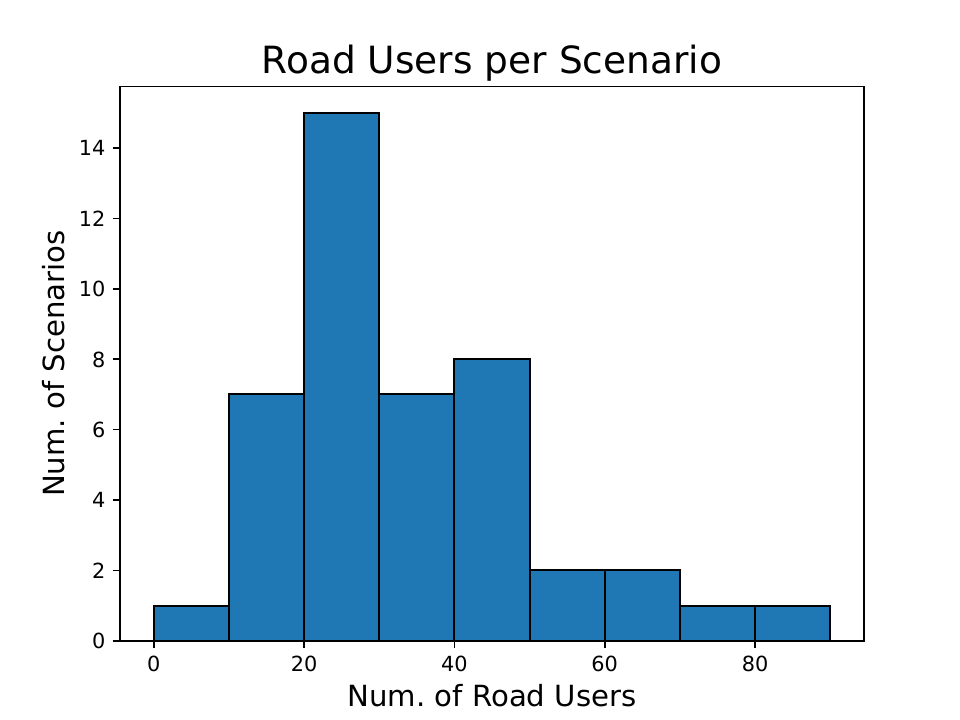}
         \caption{Traffic Distribution}
         \label{fig:traffic_distribution}
     \end{subfigure}
        \caption{Statistics of the SCOPE dataset}
        \label{fig:statistics}
        \vspace*{-3mm}
\end{figure*}

In order to achieve safe autonomous driving, algorithms must be trained and tested on data that is as comprehensive and realistic as possible. This includes not only sensor modalities, but also scenario diversity. Multiple datasets presented in Sec.~\ref{sec:related_work} only contain recordings from urban and suburban environments and neglect scenarios such as roundabouts. To address these limitations, we incorporate 44 diverse scenarios that encompass various urban areas, including small and large four-way intersections as well as T-junctions, residential areas, rural roads, and highways. However, the SCOPE dataset extends beyond the typical traffic scenarios observed in inner-city environments. It also encompasses less common scenarios, such as roundabouts and tunnel sections.
The distribution of scenarios is shown in Fig.~\ref{fig:scenario_distribution}. Approximately \SI{50}{\percent}  of the scenarios are T-junction and four-way intersection scenarios, as these types of scenarios are considered the most interesting from a collective perception standpoint. Inner-city scenarios have a high occlusion and collision potential, which makes them particularly relevant. Highway scenarios represent the third largest category, accounting for approximately \SI{16}{\percent} of the total. Highways are relevant because they represent a significant portion of natural driving and because of the high speeds, there is a high potential for CP to increase traffic safety. Additionally, on rural roads, speeds are high, and the perception range is limited due to the curvy road layout. Consequently, CP is also relevant in this context, and rural scenarios constitute \SI{9}{\percent} of the SCOPE dataset. The remaining scenarios are urban road sections without intersections, a tunnel, and roundabouts, which are also common real-world scenarios and are therefore part of the dataset.

For a higher scenario diversity, the scenarios of one category (e.g., T-junction) are distributed over several maps. The CARLA maps \textit{Town01}-\textit{Town07} and \textit{Town10} are used for this purpose.
Moreover, we propose two novel maps from Germany with a highly realistic environment modeling to better resemble real-world scenarios. The first map is a digital-twin of the autonomous driving test field in Karlsruhe. The second map is an inner-city round course in Tübingen. Example images of these two maps can be found in Fig.~\ref{fig:own_maps}. As a basis for these two maps, the corresponding OpenStreetMap~\cite{OpenStreetMap} was used and adapted for the use in CARLA with RoadRunner~\cite{RoadRunner}. Afterwards, the fast modeling approach of Schulz et al.~\cite{schulz2023} is utilized to generate static meshes of the buildings. Finally, the specific types of trees, the location of traffic signs, and the precise positioning of traffic lights were added manually. 

Figure~\ref{fig:map_distribution} shows the distribution of the scenarios on the incorporated maps.
\textit{Town03} is the most used, as it is the only map with a roundabout and a tunnel. In addition, this map contains both urban and residential areas which are both required for multiple scenarios. Many scenarios take place in \textit{Town04} because it is the only map containing a bi-directional highway. The maps of Karlsruhe, Tübingen, and \textit{Town10} have a share of \SI{13.6}{\percent} each, because they provide different urban scenarios. \textit{Town07} is used for the rural scenarios, as this map is the only one containing rural roads. The other maps (\textit{Town01}, \textit{Town02} and \textit{Town06}) are used for further residential and highway scenarios.

The diversity of scenarios is not solely contingent upon the specific scenario type and map; it is also influenced by the varying number of traffic participants and road user classes. To ensure a comprehensive evaluation, we include not only cars, vans, and motorcycles but also the VRU classes, cyclists and pedestrians, in scenarios that are reasonable, such as urban or residential areas. In terms of collaborative agents, we equip up to 20 CAVs and 4 RSUs per scenario with sensors. On average, approximately 10 collaborative agents are present. The distribution of collaborative agents is shown in Fig.~\ref{fig:agent_distribution}. Besides the sensor-equipped CAVs, passive traffic is required to simulate a realistic traffic situations with a reasonable V2X equipment rate. Hence, the scenarios in SCOPE contain up to 60 further traffic participants in the above mentioned classes as passive traffic. The highest traffic volume appears for the highway scenarios, the lowest for smaller residential area scenarios. In all scenarios \SIrange{25}{50}{\percent} of the traffic participants are CAVs.

\subsection{Sensor Setup}
\label{subsec:sensors}

\begin{figure}[t]
    \centering
    \includegraphics[width=\linewidth, trim= 1.5cm 3cm 1.5cm 1cm, clip]{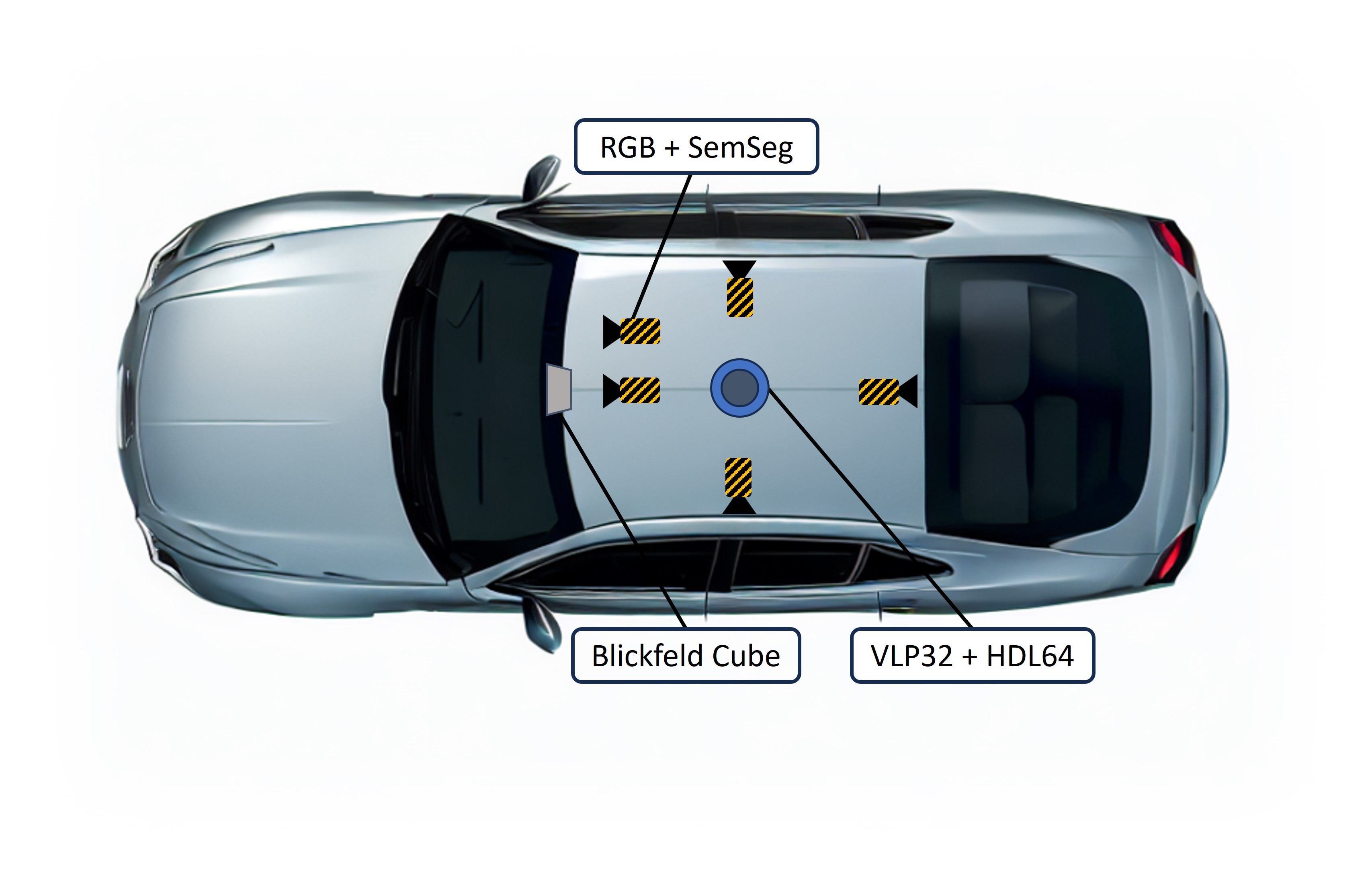}
    \caption{Sensor Setup of the CAVs} 
    \label{fig:sensor_setup}
\end{figure}

In order to ensure broad applicability, the SCOPE dataset incorporates recordings from RGB cameras, semantic segmentation (SemSeg) cameras, and three different LiDAR sensors. 
An overview of the sensor suite is shown in Fig.~\ref{fig:sensor_setup}, the specification of the sensors is presented in Tab.~\ref{tab:sensor_spec}. All sensors are configured with a recording frequency of \SI{10}{\hertz} since this corresponds to a medium frequency of the real-world LiDAR sensors and avoids too much data. 

\begin{figure*}
     \centering
     \begin{subfigure}[b]{0.49\textwidth}
         \centering
         \includegraphics[width=\textwidth, page=3, trim= 7.5cm 4cm 7.5cm 4cm, clip]{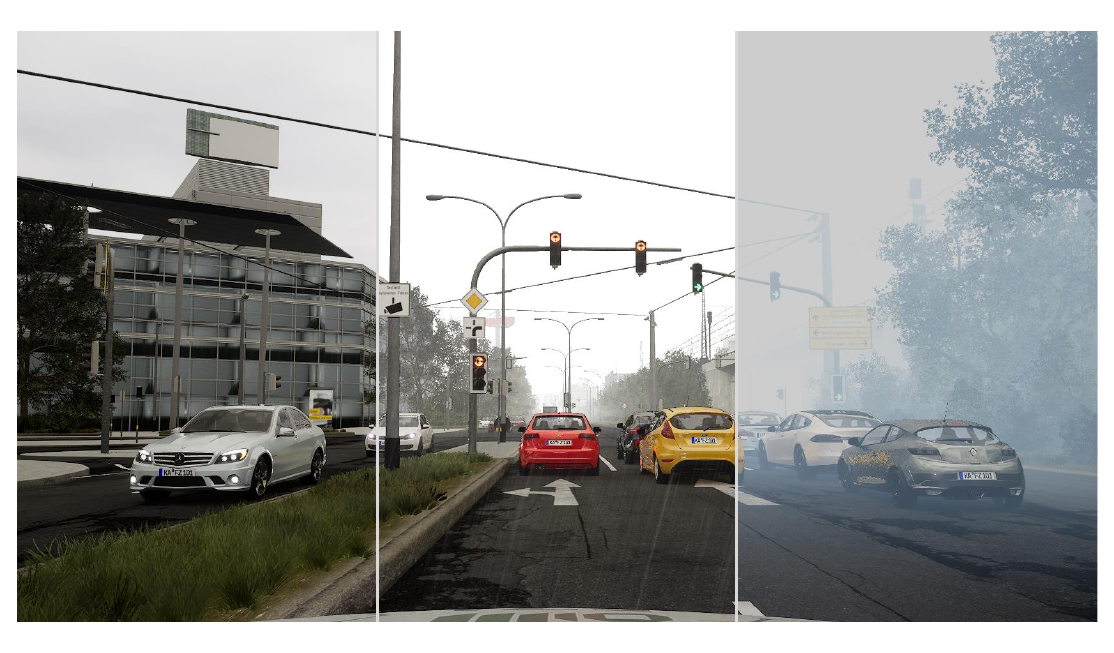}
         \caption{Camera weather simulation. Image from~\cite{CARLA}}
         \label{fig:image_weather}
     \end{subfigure}
     \hfill
     \begin{subfigure}[b]{0.49\textwidth}
         \centering
         \includegraphics[width=\textwidth, page=4, trim= 7.5cm 4cm 7.5cm 4cm, clip]{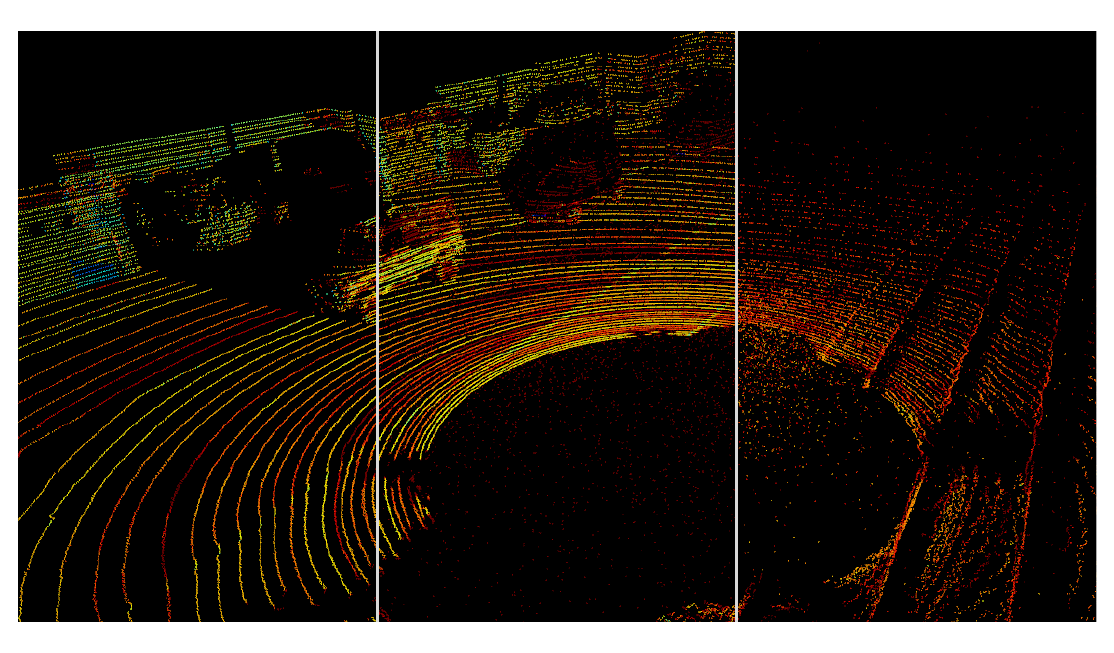}
         \caption{LiDAR weather simulation. Point Cloud from~\cite{kitti}}
         \label{fig:lidar_weather}
     \end{subfigure}
        \caption{Exemplary scenes including the weather simulation for camera data (a) and LiDAR data (b) with clear weather (left), rain (mid) and fog (right)}
        \label{fig:weather-simulations}
\end{figure*}
The camera setup consists of five RGB and five SemSeg cameras. The cameras have a resolution of 1920$\times$1080 px and a field of view (FOV) of \SI{110}{\degree}. As illustrated in Fig.~\ref{fig:sensor_setup}, two cameras are positioned in parallel with a frontal orientation, which enables the training and evaluation of monocular and stereo image object detection. The remaining three cameras are oriented to the left, right, and rear to achieve a 360° camera coverage. The resolution was chosen as it is a widely used format. The FOV of \SI{110}{\degree} aims to create an overlap of two adjacent cameras which are rotated by \SI{90}{\degree} for a better \SI{360}{\degree} surround view creation. This applies to both RGB and SemSeg cameras. In addition, to allow for a better understanding of the scenario, each vehicle is equipped with a bird's-eye view camera, which is located centered above the CAV in a height of \SI{40}{\metre}.

\begin{table}[b]
\centering
\caption{Sensor specification}
\begin{tabular}{l  l} \toprule
    Sensor & Specifications \\ \midrule
    5$\times$ RGB Camera  & Stereo + \SI{360}{\degree}, 1920$\times$1080px, \SI{110}{\degree} FOV \\
    \rule{0pt}{3ex}5$\times$ SemSeg Camera & Stereo + \SI{360}{\degree}, 1920$\times$1080px, \SI{110}{\degree} FOV \\
    \rule{0pt}{3ex}1$\times$ BEV RGB Camera & 1920$\times$1080px, \SI{110}{\degree} FOV, \SI{40}{\metre} above CAV \\
    \rule{0pt}{3ex}1$\times$ HDL64 LiDAR  & \SI{360}{\degree}$\times$\SI{26.8}{\degree} FOV, 64 Layer, \SI{10}{\hertz},\\& \SI{120}{\metre} range, 1.3M points per \si{\second} \\
    \rule{0pt}{3ex}1$\times$ VLP32 LiDAR & \SI{360}{\degree}$\times$\SI{40}{\degree} FOV, 32 Layer, \SI{10}{\hertz},\\& \SI{200}{\metre} range, 600k points per \si{\second} \\
    \rule{0pt}{3ex}1$\times$ CUBE LiDAR & \SI{72}{\degree}$\times$\SI{30}{\degree} FOV, 52 Scan Lines, \SI{10}{\hertz},\\& \SI{250}{\metre} range, 90k points per \si{\second}\\ \bottomrule
\end{tabular}
\vspace{-1mm}
\label{tab:sensor_spec}
\end{table}

The publicly available multi-modal CP datasets introduced in Sec.~\ref{sec:related_work} use the CARLA simulator~\cite{CARLA} due to the advantages described. However, the CARLA LiDAR model has some functional insufficiencies and lacks physical accuracy. 

CARLA includes a parameterizable weather simulation; however, the LiDAR sensor is not affected by this precipitation or fog. The point drop-off is only a probabilistic model and does not depend on the distance between point and sensor. Additionally, noise is solely a random shift of the point's position along the ray and no noise of the intensity is available. Furthermore, the sensor only uses the atmospheric attenuation factor and the distance to the object to calculate the intensity and neglects other effects on the intensity, such as the angle of the light beam or the material properties of the object hit. Moreover, the sensor model does not include beam divergence which is a relevant physical property of LiDAR sensors.
To overcome this domain gap while using the CARLA simulator, we incorporate the improved LiDAR sensor models proposed by Rosenberger et al.~\cite{rosenberger2020sequential}. They conducted an intensive experiment over six months and recorded the behavior of different LiDAR sensors in a real-world environment. Based on the observation they created physically-accurate sensor models for a Blickfeld CUBE, and a Velodyne VLP32. These models incorporate realistic drop-outs, noise, an improved intensity modeling and physical properties such as beam divergence and a detection threshold.

For the SCOPE dataset, we include three different LiDAR sensors as shown in Fig.~\ref{fig:sensor_setup}. A 64-layer \SI{360}{\degree} LiDAR imitating the Velodyne HDL-64 is configured using the model of the VLP32 by~\cite{rosenberger2020sequential}. This sensor achieves a range of \SI{120}{\metre} and 1.3M points per second. Additionally, we include the 32-layer \SI{360}{\degree} Velodyne VLP32 with a range of \SI{200}{\metre} to allow training and evaluation on \SI{360}{\degree} LiDAR sensors with different numbers of layers. Both \SI{360}{\degree} LiDARs are mounted centered on the roof of the vehicle. The mounting position is designed to accommodate the lowest layer, preventing it from hitting the vehicle. Furthermore, our dataset is the first to include a solid-state LiDAR (Blickfeld CUBE). As autonomous vehicles are likely to be equipped with solid-state LiDAR sensors for market readiness, it is necessary to include these in the datasets. This Blickfeld CUBE achieves a range of up to \SI{250}{\metre}. In our sensor setup, the CUBE LiDAR is configured with 52 scan lines and a FoV of \SI{72}{\degree}. The CUBE sensor is mounted at the front of the roof facing in frontal direction.

\subsection{Environmental Variation}
\label{subsec:weather}

 Environmental perception using camera or LiDAR sensors is heavily affected by environmental conditions such as rain, snow and fog. To overcome these issues, neural networks must be trained with data including these conditions~\cite{volk2019towards, Teufel-IV23}. Thus, we incorporate physically-accurate weather simulations for camera and LiDAR sensors. We include rain and fog with three intensity levels into our dataset. For rain, intensities with \SI{10}{\mm\per\hour} (low), \SI{20}{\mm\per\hour} (medium), and \SI{40}{\mm\per\hour} (high) are used. The fog intensities are \SI{0.01}{\per\micro\meter\cubed} (low), \SI{0.02}{\per\micro\meter\cubed} (medium), and \SI{0.05}{\per\micro\meter\cubed} (high). Figure~\ref{fig:weather-simulations} shows exemplary images and point clouds including the rain and fog augmentation. The distribution of the incorporated weather conditions is presented in Fig.~\ref{fig:weather_distribution}.

Weather is not the only factor that can affect camera-based object detection, varying lighting conditions can also have an effect. To address this, we have included recordings from different times of day, including periods of blinding sunlight and nighttime. In addition, these different times of day are combined with the aforementioned weather conditions to provide a comprehensive dataset also for evaluation under adverse conditions. To avoid unbalanced data, each scenario is available with sunny weather, both a daytime rain and fog augmentation, and either a nighttime scene with clear weather or nighttime with medium rain augmentation. Night and fog is not combined because this environmental condition is less likely.

\subsubsection{Image Augmentation}
\label{subsubsec:image_augmentataion}

The image augmentation with rain is conducted using the model for falling rain by Hospach et al.~\cite{fallingrain} in combination with the raindrop simulation by von Bernuth et al.~\cite{raindrops}. Exemplary scenes including the weather simulation are shown in Fig.~\ref{fig:image_weather}.

Incorporating the camera parameters and the depth map (recorded using the depth camera by CARLA~\cite{CARLA}) a 3D scene reconstruction is performed. In the reconstructed space between camera and the background falling rain streaks are rendered. The simulation respects camera parameters such as focal length, aperture, shutter speeds as well as the distance to the camera and adapts the length of the rain streaks according to these parameters. To achieve a realistic rain simulation not only falling rain but also raindrops on the windscreen/camera lens are included. The approach by von Bernuth et al.~\cite{raindrops} distributes raindrops on the surface in front of the camera lens and uses ray tracing for a physically-accurate rendering including reflections. The resulting rain simulation is highly parameterizable with varying intensities, falling angles as well as drop size, and color. 
As the approach is not capable to simulate water on the street, this effect is included by the weather model of the CARLA simulator which shows a sufficient level of detail and realism.

As second weather condition, fog is incorporated using the model by von Bernuth et al.~\cite{snow-fog}. Fog consists of little water droplets with an extremely high amount of drops compared to rain. To avoid high computation times with ray tracing, the fog model uses so called light attenuation algorithms. The fog droplets lead to a scattering and absorption of light rays in dependence to an extinction factor $\alpha$ and the distance $d$ as shown in Eq.~\eqref{eq:fog}~\cite{snow-fog}:

\begin{equation}
  \label{eq:fog}
    I = I_i e^{ - \alpha_{ext} d } + I_s (1 - e^{ - \alpha_{ext} d }),
\end{equation}
\noindent
where $I_i$ describes the pixel color of a pixel $i$ of image $I$, and $I_s$ the sky color. 

\subsubsection{Point Cloud Augmentation}
\label{subsubsec:pc_augmentation}
The LiDAR sensor weather simulation used was proposed by Teufel et al.~\cite{lidar-weather}. Point clouds in clear conditions as well as with rain and fog augmentation are shown in Fig.~\ref{fig:lidar_weather}.

The rain simulation uses raytracing on a generated rain volume. 
Multiple diverging rays are traced for each point in the point cloud with a circular pattern to simulate beam divergence. A point is modified, if its total intersection ratio is higher than a defined threshold. Depending on a second threshold, the point is moved towards the sensor or deleted, which simulates noise due to scattering and absorption of the ray by the water droplet.
For fog, the simulation applies a probabilistic model for each point to match the characteristics of fog affecting point clouds. Due to the high number of droplets in fog, a raytracing based approach would result in a high computational load. The fog simulation is parameterizable with visibility distance $v$ and further parameters as described in \cite{lidar-weather}. 


\subsection{Data Structure}

We provide all data from the sensor suite presented in Sec.~\ref{subsec:sensors}. For each scenario there are separate directories for clear weather, rain, fog and night to allow a differentiated evaluation with or without weather. Each directory, contains one directory per CAV or RSU named with the corresponding ID. All sensor data is provided together with the corresponding transformation matrices from local to world coordinate system and camera projection matrix. To gain an overview about the scenario also data from the bird's-eye view camera attached to the CAV is provided.

The ground truth is provided for each scenario framewise and includes object class, position, orientation, velocity and acceleration. The position and orientation are in 3D world coordinates. Additionally, we provide 2D bounding boxes in image coordinates for each camera. To avoid including 2D boxes of occluded objects into the training of object detectors, a filtering using the semantic segmentation images was applied.
For the semantic segmentation ground truth, the semantic classes from Cityscapes~\cite{cityscapes} are used, as this is one of the most used benchmarks for semantic segmentation.

\subsection{Toolkit}
In order to enable an easy usability, the SCOPE dataset comes with a comprehensive toolkit. The toolkit is available as python package and includes various functionalities such as for the download of different parts of the dataset. Moreover, a visualization tool for camera and LiDAR data to examine the dataset is provided. The visualization tool allows to display the point clouds together with the corresponding camera images of a selected CAV. Moreover, the ground truth bounding boxes can be displayed.
For training and testing of various object detectors and semantic segmentation models using PyTorch, the toolkit also includes a dataset class that facilitates the efficient loading of data to preprocess information.

\section{BENCHMARK}
\label{subsec:metrics}

The SCOPE dataset features a comprehensive benchmark for 2D and 3D object detection as well as semantic segmentation. We use a split of 70:10:20 which results in 12,320 frames for training, 1,760 frames for validation and 3,520 frames for testing.
The perception performance is evaluated within a range of x $\in$ [-140, 140]\si{\metre}, y $\in$ [-40, 40]\si{\metre}, z $\in$ [-4, 1]\si{\metre} around a randomly chosen ego vehicle. As evaluation metric we use the Average Precision (AP) with a 3D Intersection over Union (IoU) threshold of 0.3 (AP@IoU$_{0.3}$) and 0.5 (AP@IoU$_{0.5}$) for pedestrians and bikes and 0.5 and 0.7 (AP@IoU$_{0.7}$) for cars. The lower IoU threshold for pedestrians and bikes applies since due to the lower size of the objects perceiving them is more difficult. Additionally, we evaluate the average required bandwidth per vehicle in \si[per-mode=repeated-symbol]{\mega\bit\per\second}. Therefore, we use the size of all messages transmitted to the ego vehicle at a frequency of \SI{10}{\hertz} without communication overhead. Since the communication channel is a significant factor in collective perception, the required bandwidth used by a specific approach should also be considered. Since the communication range heavily depends on the scenario and the density within the V2X network, we do not limit the communication range. 
For the semantic segmentation benchmark we evaluate per class using the pixel-level Pascal VOC IoU (IoU class) as introduced by Everingham et al.~\cite{everingham2010pascal}. As second metric, the instance-level IoU (iIoU class) as used in the Cityscapes benchmark~\cite{cityscapes} is calculated.

\section{CONCLUSION \& OUTLOOK}
\label{sec:conclusion}

In this work, we present the novel synthetic collective perception dataset SCOPE, which includes RGB and semantic segmentation camera as well as LiDAR recordings from over 40 diverse scenarios, including edge cases such as tunnels and a roundabout for a comprehensive training and testing of collective perception algorithms. Additionally, the dataset is partially captured on two novel maps of Karlsruhe and Tübingen. The dataset features 3 to 24 collaborating agents, resulting in a total of 17,600 frames, and a wide range of V2X equipment rates. Furthermore, SCOPE is the first synthetic multi-modal dataset for collective perception that includes a realistic LiDAR model with beam divergence, as well as realistic dropouts and intensity calculations. Moreover, it is the first dataset to include varying times of day and physically-accurate weather simulation for camera and LiDAR sensors, including parameterized intensities, to improve robustness of object detection to environmental effects. Due to the wide range of sensors, camera and LiDAR-based object detectors as well as semantic segmentation algorithms can be evaluated. Since SCOPE includes different types of LiDAR sensors, the dataset is also suitable to investigate domain adaptation.

In the future, we will expand the dataset to include snowy weather conditions to cover all common weather conditions. Furthermore, we will perform an extensive evaluation using different state-of-the-art camera and LiDAR-based object detectors as well as semantic segmentation methods to create a comprehensive benchmark. All updates regarding benchmark and additional data will be published on our website.


\addtolength{\textheight}{-4.5cm}







\bibliographystyle{IEEEtran} 
\bibliography{literature.bib}

\begin{thebibliography}{10}
\providecommand{\url}[1]{#1}
\csname url@samestyle\endcsname
\providecommand{\newblock}{\relax}
\providecommand{\bibinfo}[2]{#2}
\providecommand{\BIBentrySTDinterwordspacing}{\spaceskip=0pt\relax}
\providecommand{\BIBentryALTinterwordstretchfactor}{4}
\providecommand{\BIBentryALTinterwordspacing}{\spaceskip=\fontdimen2\font plus
\BIBentryALTinterwordstretchfactor\fontdimen3\font minus \fontdimen4\font\relax}
\providecommand{\BIBforeignlanguage}[2]{{%
\expandafter\ifx\csname l@#1\endcsname\relax
\typeout{** WARNING: IEEEtran.bst: No hyphenation pattern has been}%
\typeout{** loaded for the language `#1'. Using the pattern for}%
\typeout{** the default language instead.}%
\else
\language=\csname l@#1\endcsname
\fi
#2}}
\providecommand{\BIBdecl}{\relax}
\BIBdecl

\bibitem{volk2019towards}
G.~Volk, S.~M\"uller, A.~v. Bernuth, D.~Hospach, and O.~Bringmann, ``{Towards Robust CNN-based Object Detection through Augmentation with Synthetic Rain Variations},'' in \emph{2019 IEEE Intelligent Transportation Systems Conference (ITSC)}, Oct 2019.

\bibitem{Teufel-IV23}
S.~Teufel, J.~Gamerdinger, G.~Volk, C.~Gerum, and O.~Bringmann, ``Enhancing robustness of {LiDAR-Based} perception in adverse weather using point cloud augmentations,'' in \emph{2023 IEEE Intelligent Vehicles Symposium (IV) (IEEE IV 2023)}, Anchorage, USA, Jun. 2023.

\bibitem{volk_environment-aware_2019}
G.~Volk, A.~von Bemuth, and O.~Bringmann, ``\BIBforeignlanguage{en}{Environment-aware {Development} of {Robust} {Vision}-based {Cooperative} {Perception} {Systems}},'' in \emph{\BIBforeignlanguage{en}{2019 {IEEE} {Intelligent} {Vehicles} {Symposium} ({IV})}}.\hskip 1em plus 0.5em minus 0.4em\relax Paris, France: IEEE, Jun. 2019.

\bibitem{schiegg2021}
F.~A. Schiegg, I.~Llatser, D.~Bischoff, and G.~Volk, ``Collective perception: A safety perspective,'' \emph{Sensors}, vol.~21, no.~1, 2021.

\bibitem{gamerdinger2023cold}
J.~Gamerdinger, S.~Teufel, G.~Volk, and O.~Bringmann, ``Cold fusion: A real-time capable spline-based fusion algorithm for collective lane detection,'' in \emph{2023 IEEE Intelligent Vehicles Symposium (IV)}.\hskip 1em plus 0.5em minus 0.4em\relax IEEE, 2023.

\bibitem{teufel2023collective}
S.~Teufel, J.~Gamerdinger, G.~Volk, and O.~Bringmann, ``Collective pv-rcnn: A novel fusion technique using collective detections for enhanced local lidar-based perception,'' in \emph{2023 IEEE 26th International Conference on Intelligent Transportation Systems (ITSC)}.\hskip 1em plus 0.5em minus 0.4em\relax IEEE, 2023.

\bibitem{kitti}
A.~Geiger, P.~Lenz, and R.~Urtasun, ``Are we ready for autonomous driving? the kitti vision benchmark suite,'' in \emph{Conference on Computer Vision and Pattern Recognition (CVPR)}, 2012.

\bibitem{waymo}
P.~Sun, H.~Kretzschmar, X.~Dotiwalla, A.~Chouard, V.~Patnaik, P.~Tsui, J.~Guo, Y.~Zhou, Y.~Chai, B.~Caine, V.~Vasudevan, W.~Han, J.~Ngiam, H.~Zhao, A.~Timofeev, S.~Ettinger, M.~Krivokon, A.~Gao, A.~Joshi, Y.~Zhang, J.~Shlens, Z.~Chen, and D.~Anguelov, ``Scalability in perception for autonomous driving: Waymo open dataset,'' in \emph{IEEE/CVF Conference on Computer Vision and Pattern Recognition}, June 2020.

\bibitem{teufel2024review}
S.~Teufel, J.~Gamerdinger, J.-P. Kirchner, G.~Volk, and O.~Bringmann, ``Collective perception datasets for autonomous driving: A comprehensive review,'' in \emph{2024 IEEE Intelligent Vehicles Symposium (IV)}.\hskip 1em plus 0.5em minus 0.4em\relax IEEE, 2024.

\bibitem{rosenberger2020sequential}
P.~Rosenberger, M.~F. Holder, N.~Cianciaruso, P.~Aust, J.~F. Tamm-Morschel, C.~Linnhoff, and H.~Winner, ``Sequential lidar sensor system simulation: a modular approach for simulation-based safety validation of automated driving,'' \emph{Automotive and Engine Technology}, vol.~5, 2020.

\bibitem{CARLA}
A.~Dosovitskiy, G.~Ros, F.~Codevilla, A.~Lopez, and V.~Koltun, ``{CARLA: An Open Urban Driving Simulator},'' in \emph{Conference on robot learning}.\hskip 1em plus 0.5em minus 0.4em\relax PMLR, 2017.

\bibitem{wang2020v2vnet}
T.-H. Wang, S.~Manivasagam, M.~Liang, B.~Yang, W.~Zeng, and R.~Urtasun, ``V2vnet: Vehicle-to-vehicle communication for joint perception and prediction,'' in \emph{Computer Vision--ECCV 2020: 16th European Conference, Glasgow, UK, Proceedings, Part II 16}.\hskip 1em plus 0.5em minus 0.4em\relax Springer.

\bibitem{codd}
E.~Arnold, S.~Mozaffari, and M.~Dianati, ``Fast and robust registration of partially overlapping point clouds,'' \emph{IEEE Robotics and Automation Letters}, 2021.

\bibitem{comap}
Y.~Yuan and M.~Sester, ``Comap: A synthetic dataset for collective multi-agent perception of autonomous driving,'' \emph{The International Archives of the Photogrammetry, Remote Sensing and Spatial Information Sciences}, vol. XLIII-B2-2021, 2021.

\bibitem{xu2022v2xvit}
R.~Xu, H.~Xiang, Z.~Tu, X.~Xia, M.-H. Yang, and J.~Ma, ``V2x-vit: Vehicle-to-everything cooperative perception with vision transformer,'' in \emph{European conference on computer vision}.\hskip 1em plus 0.5em minus 0.4em\relax Springer, 2022.

\bibitem{dolphins}
R.~Mao, J.~Guo, Y.~Jia, Y.~Sun, S.~Zhou, and Z.~Niu, ``Dolphins: Dataset for collaborative perception enabled harmonious and interconnected self-driving,'' in \emph{Proceedings of the Asian Conference on Computer Vision}, 2022.

\bibitem{opv2v}
R.~Xu, H.~Xiang, X.~Xia, X.~Han, J.~Li, and J.~Ma, ``Opv2v: An open benchmark dataset and fusion pipeline for perception with vehicle-to-vehicle communication,'' in \emph{2022 International Conference on Robotics and Automation (ICRA)}.\hskip 1em plus 0.5em minus 0.4em\relax IEEE, 2022.

\bibitem{V2XSim}
Y.~Li, D.~Ma, Z.~An, Z.~Wang, Y.~Zhong, S.~Chen, and C.~Feng, ``V2x-sim: Multi-agent collaborative perception dataset and benchmark for autonomous driving,'' \emph{IEEE Robotics and Automation Letters}, 2022.

\bibitem{wei2023asynchrony}
S.~Wei, Y.~Wei, Y.~Hu, Y.~Lu, Y.~Zhong, S.~Chen, and Y.~Zhang, ``Asynchrony-robust collaborative perception via bird's eye view flow,'' \emph{arXiv e-prints}, 2023.

\bibitem{manivasagam2020lidarsim}
S.~Manivasagam, S.~Wang, K.~Wong, W.~Zeng, M.~Sazanovich, S.~Tan, B.~Yang, W.-C. Ma, and R.~Urtasun, ``Lidarsim: Realistic lidar simulation by leveraging the real world,'' in \emph{Proceedings of the IEEE/CVF Conference on Computer Vision and Pattern Recognition}, 2020, pp. 11\,167--11\,176.

\bibitem{xu2021opencda}
R.~Xu, Y.~Guo, X.~Han, X.~Xia, H.~Xiang, and J.~Ma, ``Opencda: an open cooperative driving automation framework integrated with co-simulation,'' in \emph{2021 IEEE International Intelligent Transportation Systems Conference (ITSC)}.\hskip 1em plus 0.5em minus 0.4em\relax IEEE, 2021.

\bibitem{SUMO}
P.~A. Lopez, M.~Behrisch, L.~Bieker-Walz, J.~Erdmann, Y.-P. Flötteröd, R.~Hilbrich, L.~Lücken, J.~Rummel, P.~Wagner, and E.~Wiessner, ``Microscopic traffic simulation using sumo,'' in \emph{2018 21st International Conference on Intelligent Transportation Systems (ITSC)}, 2018.

\bibitem{resist}
S.~Mueller, D.~Hospach, J.~Gerlach, O.~Bringmann, and W.~Rosenstiel, ``{Framework for Varied Sensor Perception in Virtual Prototypes},'' in \emph{MBMV}, 2015.

\bibitem{lidar-weather}
S.~Teufel, G.~Volk, A.~Von~Bernuth, and O.~Bringmann, ``{Simulating Realistic Rain, Snow, and Fog Variations For Comprehensive Performance Characterization of LiDAR Perception},'' in \emph{2022 IEEE 95th Vehicular Technology Conference: (VTC2022-Spring)}, 2022.

\bibitem{OpenStreetMap}
{OpenStreetMap contributors}, ``{Planet dump retrieved from https://planet.osm.org },'' https://www.openstreetmap.org, 2017.

\bibitem{RoadRunner}
{MathWorks}, ``{RoadRunner },'' \\https://de.mathworks.com/products/roadrunner.html.

\bibitem{schulz2023}
P.~Schulz, A.~Viehl, and O.~Bringmann, ``Fast environment generation methods for virtual testing,'' in \emph{2023 IEEE 26th International Conference on Intelligent Transportation Systems (ITSC)}, 2023.

\bibitem{fallingrain}
D.~Hospach, S.~Mueller, W.~Rosenstiel, and O.~Bringmann, ``{Simulation of Falling Rain for Robustness Testing of Video-Based Surround Sensing Systems},'' in \emph{Proceedings of the 2016 Design, Automation {\&} Test in Europe Conference {\&} Exhibition (DATE)}, 2016.

\bibitem{raindrops}
A.~von Bernuth, G.~Volk, and O.~Bringmann, ``{Rendering Physically Correct Raindrops on Windshields for Robustness Verification of Camera-based Object Recognition},'' in \emph{2018 IEEE Intelligent Vehicles Symposium (IV)}, 2018.

\bibitem{snow-fog}
A.~v. Bernuth, G.~Volk, and O.~Bringmann, ``{Simulating Photo-realistic Snow and Fog on Existing Images for Enhanced CNN Training and Evaluation},'' in \emph{2019 IEEE Intelligent Transportation Systems Conference (ITSC)}, 2019.

\bibitem{cityscapes}
M.~Cordts, M.~Omran, S.~Ramos, T.~Rehfeld, M.~Enzweiler, R.~Benenson, U.~Franke, S.~Roth, and B.~Schiele, ``The cityscapes dataset for semantic urban scene understanding,'' in \emph{Proc. of the IEEE Conference on Computer Vision and Pattern Recognition (CVPR)}, 2016.

\bibitem{everingham2010pascal}
M.~Everingham, L.~Van~Gool, C.~K. Williams, J.~Winn, and A.~Zisserman, ``The pascal visual object classes (voc) challenge,'' \emph{International journal of computer vision}, vol.~88, pp. 303--338, 2010.

\end{thebibliography}

\end{document}